\documentclass{article}

\usepackage[final]{nips_2017}
\usepackage{graphicx}

\usepackage[utf8]{inputenc} % allow utf-8 input
\usepackage[T1]{fontenc}    % use 8-bit T1 fonts
\usepackage{hyperref}       % hyperlinks
\usepackage{url}            % simple URL typesetting
\usepackage{booktabs}       % professional-quality tables
\usepackage{amsfonts}       % blackboard math symbols
\usepackage{nicefrac}       % compact symbols for 1/2, etc.
\usepackage{microtype}      % microtypography
\usepackage{mathtools}

\title{Convolutional Neural Networks for Breast Cancer Screening:Transfer Learning with Exponential Decay}

\author{
  Hiba Chougrad \\
    Laboratoire d'Informatique et Mathématiques\\ et leurs Applications (LIMA), Faculty of science\\
  Chouaib Doukkali University\\
  Morocco, El JADIDA 24000 \\
  \texttt{chougrad.hiba@gmail.com} \\
   \And
   Hamid Zouaki \\
   LIMA,Faculty of science  \\
   Chouaib Doukkali University\\
   \texttt{hamid\_zouaki@yahoo.fr} \\
   \AND
   Omar Alheyane \\
   LMF, Faculty of science \\
   Chouaib Doukkali University\\
   \texttt{alehyane@hotmail.com} \\
}

\begin{document}

\maketitle

\begin{abstract}
In this paper, we propose a Computer Assisted Diagnosis (CAD) system based on a deep Convolutional Neural Network (CNN) model, to build an end-to-end learning process that classifies breast mass lesions. We investigate the impact that has transfer learning when large data is scarce, and explore the proper way to fine-tune the layers to learn features that are more specific to the new data. The proposed approach showed better performance compared to other proposals that classified the same dataset.
\end{abstract}

\section{Background and objectives}

Breast cancer is the most common invasive disease among women \citep{siegel2014cancer}
Optimistically, an early diagnosis of the disease increases the chances of recovery dramatically and as such, makes the early detection crucial. Mammography is the recommended screening technique, but it is not enough, we also need the radiologist expertise to check the mammograms for lesions and give a diagnosis, which can be a very challenging task\citep{kerlikowske2000performance}. 
Radiologists often resort to biopsies and this ends up adding exorbitant expenses to an already burdened patient and health care system \citep{sickles1991periodic}. 
We propose a Computer Assisted Diagnosis (CAD) system, based on a deep Convolutional Neural Network (CNN) model, designed to be used as a “second-opinion” to help the radiologist give more accurate diagnoses. Deep Learning requires large datasets to train networks of a certain depth from scratch, which are lacking in the medical domain especially for breast cancer. Transfer learning proved to be efficient to deal with little data, even if the knowledge transfer is between two very different domains \citep{shin2016deep}. 
But still using the technique can be tricky, especially with medical datasets that tend to be unbalanced and limited. And when using the state-of-the art CNNs which are very deep, the models are highly inclined to suffer from overfitting even with the use of many tricks like data augmentation, regularization and dropout.
The number of layers to fine-tune and the optimization strategy play a substantial role on the overall performance  \citep{yosinski2014transferable}. This raises few questions:
\begin{itemize}
\item Is Transfer Learning really beneficial for this application?
\item How can we avoid overfitting with our small dataset ?
\item How much fine-tuning do we need? and what is the proper way to do it?
\end{itemize}
We investigate the proper way to perform transfer learning and fine-tuning, which will allow us to take advantage of the pre-trained weights and adapt them to our task of interest. We empirically analyze the impact of the fine-tuned fraction on the final results, and we propose to use an exponentially decaying learning rate to customize all the pre-trained weights from ImageNet\citep{deng2009imagenet} and make them more suited to our type of data. The best model can be used as a baseline to predict if a new “never-seen” breast mass lesion is benign or malignant.

\section{Methodology}

We use BCDR-F03 dataset \citep{lopez2012bcdr}from the Breast Cancer Digital Repository (BCDR-FM) which contains digitized mammograms from 344 patients. It is a binary class dataset composed of benign and malignant findings. To get a balanced dataset, we used a subset of the data comprised of 300 cases of patients which gave us 600 images, 300 images showing benign lesions and 300 showing malignant lesions. We can later use the remaining cases of image lesions, from the dataset, for prediction to test the performance of the proposed CAD system, since they can be considered as “never-seen” images.

Each patient’s case includes mammograms with the associated coordinates of the lesions’ contours.
We followed the same pre-processing steps of \citep{arevalo2016representation} and \citep{perre2017lesion} except for the size of the cropped ROIs (Regions Of  Interest) which were 299 x 299 instead of 150 x 150,  so that the resulting image lesions are compatible with the ImageNet images that were used to train the original Inception-v3 CNN model. 
All the images were preprocessed as follows: 
\begin{enumerate}
\item First we cropped fixed sized regions of interest using the ground truth provided with the dataset.\label{item:1}
\item Then, we used global contrast normalization where every image $\mathbf{x}_{i,j}$ was normalized by subtracting the mean and dividing by the standard deviation of its elements.\label{item:2}
\item Then we used data augmentation by applying series of random transformations to the images i.e. width and height shifts by a fraction of 0.25, random rotation that ranges from 0-40 degrees and random horizontal flip. \label{item:3} 
\end{enumerate}

Transfer learning and fine-tuning aim at reusing the pre-trained weights of a CNN as an initialization for a new task of interest. The CNN model Inception-v3 \citep{szegedy2016rethinking} devised a new module named "The inception module" which is a 4 parallel pathway of 1x1, 3x3 and 5x5 convolution filters. The architecture of this module allows the model to recover both local features via smaller convolutions and high abstracted features via large ones. And because of the parallel network implementation in addition to the down sampling layers in each block, the model's execution time beats other state-of-the-art CNNs.

\begin{figure}[h]
  \centering
  \includegraphics[width=\linewidth]{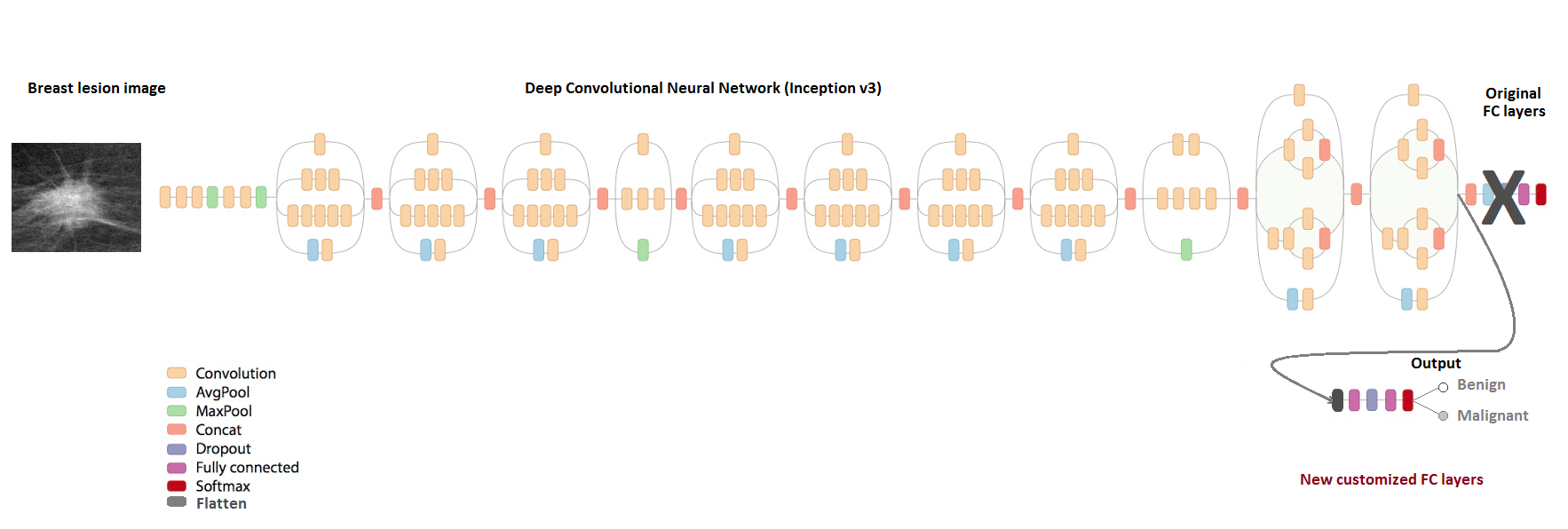}
  \caption{The customized inception v3}
  \label{fig:CNN}
\end{figure}

The Inception-v3 CNN consists of two parts:
(1) The convolutional neural network part for feature extraction and (2) the classification part with the fully-connected and softmax layers. We removed the fully-connected part of the model and built one customized to our number of classes (i.e. Benign and Malignant. See figure ~\ref{fig:CNN}).
The resulting model is composed of 221 layers. The top convolutional part, which we kept unchanged, has 5 convolutional layers  each one followed by batch normalization, 2 pooling layers and 11 inception modules. While the customized fully-connected part of the model we added, is composed of a flatten layer, 2 fully-connected dense layers and a dropout layer to randomly turn off the activations at training time with a probability of 0.5. Finally, the softmax layer is used to give normalized class probabilities for the output being "Benign” or "Malignant".  
We conducted four main experiments:
\begin{itemize}
\item (i) Inceptionv3-RI: we used randomly initialized weights vs pre-trained weights. 
\item (ii) Inceptionv3-FE: we used the pre-trained network as a fixed feature extractor.
\item (iii) Inceptionv3-1FT, Inceptionv3-2FT, Inceptionv3-3FT, Inceptionv3-AllFT: we fine-tuned the weights of the pre-trained network by gradually unfreezing the bottom layers of the convolutional part of the model (the last block, the last two blocks, the last three blocks and all the convolutional blocks).
\item (iv) Inceptionv3-FTED: we fine-tuned all the weights by resuming back-propagation on all the layers with a per-layer exponentially decaying learning rate.
\end{itemize}

According to \citep{yosinski2014transferable}, the first layers of a CNN learn generic features, while the last layers tend to be more specific to the data. The assumption is that when we froze all top layers and fine-tuned only the last two, we helped the model learn more data-specific features. 

In other words, when fine-tuning, the weights of the last convolutional layers need to be tweaked as much as possible, while the ones from the top layers remain nearly untouched. 
In accordance with this, we propose to fine-tune all the layers of the network using a per-layer exponentially decaying learning rate  \citep{walter1990nonlinear} (Inceptionv3-FTED).
We set the learning rate to be a bit high (1E-3) for the last convolutional layers and we decrease its value gradually per-layer in an exponentially-decaying manner as we move backwards to the top layers. 
We gradually update the layers starting from the last ones going towards the top layers with a corresponding reduction in the learning rate.
The learning rate $\mathbf t_{l}^{(k)} $ decays exponentially such as:
$\mathbf t_{l}^{(k)}= t_{0} \ast \exp (-\lambda \ast l)$ with $\mathbf \lambda $ a hyperparameter set to -3 and $\mathit{l}$ the number of layers. The update equation for SGD with momentum is then: $\mathbf \Delta x_{l}^{(k)}= \mu \Delta x_{l}^{(k-1)}-t_{l}^{(k)} g^{(k)} $ 
The update rule of the gradient descent searches the direction given the negative gradient $\mathbf g^{(k)} $. $\mathbf \Delta x_{l}^{(k)} $ denotes the update in the parameters of the $\mathit{l}$ -th layer and $\mathit{k}$ -th epoch and $\mathbf \mu $ is the momentum coefficient.   

\section{Results and discussion}

Experiments were run 5 times using 80\% of the data as training/validation data and 20\% as testing data. We trained the models for 90 epochs with a batch size of 128. We used Adam optimizer to train from scratch and SGD with momentum of 0.9 when fine-tuning , with a learning rate set to 1E-4; which was then divided by 10 each time the validation error stopped improving. The initial learning rate for Inceptionv3-FTED is 1E-3 which will be annealed at each layer. We adopted an early stopping strategy to monitor the validation loss with a patience set to 15 epochs. We also used data augmentation, regularization and dropout to avoid overfitting. 

\begin{figure}
  \centering
  \includegraphics[scale=0.6]{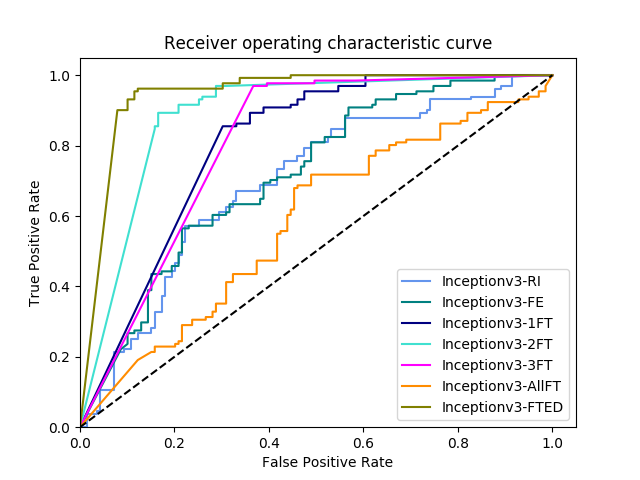}
  \caption{The ROC curve plot of the classification of the test data using the model variants.}
  \label{fig:roc}
\end{figure}

Table~\ref{res} gives the mean accuracy, standard deviation and time elapsed for each model and figure~\ref{fig:roc} gives the ROC curve plot of the classification of the test data by the model variants.  

\begin{table}[h]
  \caption{Comparison summary between the different model implementations.}
  \label{res}
  \centering
  \begin{tabular}{llll}
    \toprule          
    Model    & Accuracy(\%)   & Std(\%)  &Time elapsed(min) \\
    \midrule
    Inceptionv3-RI   	& 75.83 	& $\pm$1.53    &176.2 		\\
    Inceptionv3-FE   	& 92.67 	& $\pm$1.79    &59.1		\\
    Inceptionv3-1FT  	& 94.17 	& $\pm$0.92    &61.6		\\
    Inceptionv3-2FT  	& 96.67 	& $\pm$0.85    &63.9		\\
    Inceptionv3-3FT  	& 95.17 	& $\pm$1.18    &67.1		\\
    Inceptionv3-AllFT   & 92.67	& $\pm$1.34    &158.9		\\
    Inceptionv3-FTED    & \textbf{97.50}	& $\pm$1.26   &161.5		\\
    \bottomrule
  \end{tabular}
\end{table}

Experiments showed that transfer learning is advantageous to our task of interest and this in terms of time and accuracy (Inceptionv3-RI vs Inceptionv3-FE). Even if the transfer is between two unrelated tasks, the use of pre-trained weights for initialization is a good way to start the learning process. The next step is to start fine-tuning the loaded weights so that we train the network to better fit the new dataset. This is achieved by resuming backpropagation of the layers with a small learning rate. We gradually unfroze the last convolutional blocks of the model and tested the different model variants on the data. The performance increased when we started fine-tuning the last convolutional layers up until 2 convolutional blocks (Inceptionv3-2FT), when 3 or more blocks were fine-tuned the accuracy dropped. The idea here is that we need to make the last convolutional layers learn more data-specific features, but there is no need for us to change the weights of the first convolutional layers as much; they're already well-tuned to learn generic features especially if we don't have enough data to train on. The middle convolutional layers might be responsible of learning somewhat more complex features compared to the first layers and thus they need to be slightly modified. The exponentially decaying learning rate helps us in this regard.

The proposed method for fine-tuning (Inceptionv3-FTED) showed better performance compared to other proposals that classified the same dataset(see table~\ref{comp}).
The per-layer decaying learning rate helps us control the rate at which weights change for each part of the network. When fine-tuning, we want some of the layers to be more or less receptive to change. Transfer learning with a per-layer exponentially decaying learning rate yielded better results. 
The model Inceptionv3-FTED demonstrated an accuracy of 97.50\% and an AUC of 0.96 , thus outperforming human-level performance \citep{elmore2009variability} and hence the model can be used as a baseline to build a powerful tool for assisting radiologists in improving the efficiency, consistency, and accuracy of breast cancer diagnosis.

\begin{table}[h]
  \caption{Comparison with other proposals that classified the same datastet.}
  \label{comp}
  \centering
  \begin{tabular}{lll}
    \toprule          
    Model    				    & Accuracy(\%)  & AUC(Area Under the Curve)   \\
    \midrule
    \citep{perez2014improving}  		& -  		&0.8562     \\
    \citep{arevalo2016representation}& -  		&0.826      \\
    \citep{padillabreast}   		 	&-  			&0.950      \\
    \citep{jadoon2017three}          &83.74\%  	&-    	    \\
    \citep{perre2017lesion}   	    &-  			&0.793    	\\
     Ours   				& \textbf{97.50}			&\textbf{0.96}        \\
    \bottomrule
  \end{tabular}
\end{table}

\subsubsection*{Acknowledgments}
The authors would like to thank Stephan Gouws and Nyalleng Moroosi for their valuable and constructive suggestions during the presentation of the first draft of this research at "Deep Learning Indaba 2017".\\ 
The database used in this work was a Courtesy of MA Guevara Lopez and coauthors, Breast Cancer Digital Repository Consortium.

\bibliographystyle{plainnat}        
\bibliography{References}

\end{document}